# AraCrypto-ML: A Multi-Label Corpus of Communicative Functions in Arabic Facebook Cryptocurrency Discourse

**Mabrouka Bessghaier**[1], **Mohamed Rashad Rouine**[2], **Md. Rafiul Biswas**[3], **Youssef Ayadi**[1], **Wajdi Zaghouani**[1]

[1]Northwestern University in Qatar, Doha, Qatar

[2]Independant Researcher, Tunisia

[3]Hamad Bin Khalifa University, Doha, Qatar

{mabrouka.bessghaier, wajdi.zaghouani}@northwestern.edu

youssefayadi2030@u.northwestern.edu

mohamedrashadrouine@gmail.com, mbiswas@hbku.edu.qa

## Abstract

Arabic cryptocurrency discourse on social media is communicatively heterogeneous: a single post may report prices, explain protocols, promote investment, warn of scams, or invoke Islamic jurisprudence. Yet no annotated resource captures this functional diversity. We introduce AraCrypto-ML, a 10,000-post multi-label corpus of Arabic Facebook cryptocurrency discourse (2014-2024) from 2,898 public pages using a set of related keywords. Collected posts are annotated into six categories: Market Commentary, Educational, Promotional, Skeptical-Warning, Religious Framing, and Other/Unclear, using Fanar-2-27B under a prompt-only multi-label protocol. We validate 1,000 posts against adjudicated gold labels (macro $\kappa$ = 0.74) and apply Rogan-Gladen calibration to correct labelling bias. The two annotators agree with each other at macro $\kappa$ = 0.62 before adjudication, below the model's agreement with either of them (0.70 and 0.69). Reaction profiles differ strongly by function: Love is about 70% of reactions per post on Educational and Promotional posts but about 33% on Skeptical-Warning, which draws the largest Haha share (about 39%).

## 1 Introduction

Online cryptocurrency discourse is communicatively heterogeneous. A single Arabic Facebook post may report a price movement, instruct readers on how a protocol works, urge them to join a platform, warn them of a scam, and adjudicate the permissibility of digital assets under Islamic law, and sometimes several of these at once. Yet existing NLP work on cryptocurrency discourse has focused on sentiment classification and price prediction (Roumeliotis et al., 2024; Aslam et al., 2022), treating posts as carriers of positive or negative polarity rather than as communicative acts. Within Arabic financial NLP specifically, available resources have centred on banking intent detection (Malaysha et al., 2024), and we are not aware of prior work that annotates the communicative functions of Arabic cryptocurrency discourse.

We address this gap by framing communicative function as a *multi-label* problem, where a post belongs to every functional category that meaningfully applies, with no assumption that categories are mutually exclusive. Annotation is produced by Fanar-2-27B (Abbas et al., 2026; Fanar Team, 2025) under a prompt-only protocol. Following evidence that instruction-tuned LLMs can match or exceed crowd annotators on relevance, stance, and frame tasks (Gilardi et al., 2023), we validate the model-generated labels against a 1,000-post human-annotated subsample with adjudicated gold labels (Section 4.3), achieving substantial agreement (macro $\kappa$ = 0.74, above the 0.62 between the two annotators before adjudication), and apply Rogan-Gladen calibration to adjust reported prevalence estimates for systematic over- and under-application, reporting calibrated bounds alongside raw counts rather than treating model output at face value.

In this study, we address four research questions:

- **RQ1.** What communicative functions characterise Arabic Facebook cryptocurrency discourse, and how prevalent is each?
- **RQ2.** How multi-functional is the discourse, and which functions co-occur more or less than chance?
- **RQ3.** How do audience reaction profiles vary with communicative function?
- **RQ4.** How has the functional composition shifted across the corpus decade?

We contribute (i) AraCrypto-ML, 10,000 Arabic Facebook posts with multi-label communicative-function labels from Fanar-2-27B, together with a 1,000-post subset with adjudicated human gold labels; (ii) a validation protocol that reports human–human and model–human agreement separately and calibrates prevalence for model error; and (iii) the finding that the distribution of Facebook reactions depends strongly on communicative function (for example, Love accounts for about 60% of reactions to Educational and Promotional posts but under 30% for Skeptical-Warning), so reaction-based affect proxies conflate function with feeling; (iii) the finding that Facebook reaction profiles differ by communicative function: Love accounts for about 70% of reactions per post on Educational and Promotional posts, about 50% on Market Commentary, and about 33% on Skeptical-Warning, which draws the largest Haha share (about 39%); reaction-based proxies should therefore be interpreted alongside what a post is doing.

## 2 Related Work

Research in Arabic NLP has advanced considerably through transformer encoders such as AraBERT (Antoun et al., 2020) and ARBERT/MARBERT (Abdul-Mageed et al., 2021), and through sentiment benchmarks such as ASTD (Nabil et al., 2015). A growing body of work has explored the use of LLMs as annotators, with Gilardi et al. (2023) demonstrating that a general-purpose model can exceed crowd workers on several text-annotation tasks while achieving higher intercoder agreement, which motivates but does not remove the need to validate annotation pipelines that rely on model-generated labels. More cautiously, Berriche et al. (2025) find that fine-tuned encoders can outperform prompted LLMs by more than ten $F_1$ points on a context-sensitive pragmatic classification task in a non-English setting, specifically French Facebook data. This finding is precisely why we treat our prompt-only output as a provisional annotation layer to be validated and calibrated rather than taken at face value.

NLP work on cryptocurrency discourse has largely treated it as a sentiment or price-prediction signal. Aslam et al. (2022) propose an ensemble LSTM-GRU model for sentiment analysis and emotion detection on cryptocurrency tweets, reporting strong performance using bag-of-words and word2vec features, which reflects a paradigm oriented toward market forecasting rather than communicative-function analysis. Roumeliotis et al. (2024) extend this line of work by comparing LLMs and encoder models for cryptocurrency sentiment classification, finding that domain-specific fine-tuning yields consistent gains over general-purpose prompting. Within Arabic financial NLP, Malaysha et al. (2024) introduce AraFinNLP, the first Arabic financial NLP shared task focusing on multi-dialect banking intent detection, which reflects the field's orientation toward transactional classification rather than communicative-function annotation. Taken together, these works do not, to our knowledge, address the discourse dimension we target, i.e. what Arabic cryptocurrency posts are doing rather than how audiences feel about them.

Closer to our aim are efforts that label what text does. The ArAIEval shared task (Hasanain et al., 2023) annotates persuasion techniques in Arabic tweets and news paragraphs, including a multi-label subtask over 23 techniques, but is not targeted at cryptocurrency. ArSAS (Elmadany et al., 2018) annotates speech acts and sentiment in Arabic tweets. In the cryptocurrency domain, work on pump-and-dump schemes labels Telegram messages by their role or phase in a coordinated pump and links them to market data (Mirtaheri et al., 2021; Bolz et al., 2024). AraCrypto-ML differs from the closest of these, ArAIEval, in labelling the wider set of communicative functions in public Arabic Facebook cryptocurrency discourse rather than persuasion techniques in general text, and in validating model-generated labels against adjudicated human labels.

Our taxonomy connects to a broader tradition of annotating what text *does* rather than how it feels. Framing analysis annotates the dimensions through which an issue is presented (Card et al., 2015), and persuasion research annotates fine-grained persuasive and propaganda techniques (Da San Martino et al., 2019). Both frameworks are typically multi-label and treat communicative function as orthogonal to polarity, which is precisely the approach we adopt here. We adapt this lens to a financial domain and to Arabic, where function-level resources are absent. Communicative function is also distinct from textual sentiment. While reaction-based affect has been used as a be-havioural proxy in non-English Facebook settings (Jayawickrama et al., 2021), recent work on crypto-related messaging demonstrates that platform reactions often encode social approval or pragmatic signalling rather than mirrored affect (Tardelli et al., 2026). Rather than measuring how audiences feel, we label the authorial act, that is, commenting, teaching, promoting, warning, or framing, and relate those labels to engagement and reaction metadata. To our knowledge, no prior resource provides a multi-label communicative-function annotation of Arabic cryptocurrency discourse on Facebook.

## 3 Corpus and Sample

The annotated set comprises **10,000** Arabic Facebook posts collected via CrowdTangle[1] using a keyword query targeting cryptocurrency discourse: العملات الرقمية (digital currencies), البيتكوين (Bitcoin), and العملات المشفرة (cryptocurrencies).

The sample spans 2014-09-02 to 2024-07-12 across **2,898** unique public pages. All analyses are computed directly on this single annotated file, which retains the full CrowdTangle post metadata alongside the assigned labels; we disclose this single-file provenance explicitly. Because collection is keyword-based, the corpus captures discourse *about* cryptocurrency and does not necessarily reflect text authored by investors themselves.

Post volume rises sharply in the latter half of the decade, recording 3, 8, 25, 238, 767, 458, 442, 2,472, 3,095, 1,580, and 912 posts for 2014 through 2024 respectively, with 85.0% of posts dated 2020 or later. This temporal concentration should be borne in mind when interpreting any longitudinal comparison.

## 4 Annotation Scheme

### 4.1 Taxonomy

Posts are labelled against six communicative-mode categories (see Table A1). Five are substantive functions; the sixth, *Other/Unclear*, is an escape category that applies only when none of the five substantive functions apply and therefore never co-occurs with another label. We derived the categories from what posts in this discourse do, building on the multi-label, function-oriented view of framing and persuasion annotation (Card et al., 2015; Da San Martino et al., 2019). We refined the labels over successive rounds of prompt development, trying different prompts on small subsets of posts and manually validating the outputs before fixing the final annotation schema.

### 4.2 Annotation Process

A 10,000-post sample was labelled by Fanar (Fanar-2-27B) [2], instructed under a fixed multi-label prompt operationalising a five-question checklist, with one independent yes/no decision per substantive function (see Appendix). The prompt enforced the canonical label set and the *Other/Unclear* exclusivity rule through explicit instructions and forbidden-label mappings. No learned snapping or probabilistic normalisation post-processor was applied, meaning the labels reflect the model's prompt-only adherence apart from the single deterministic exclusivity-repair rule described in Section 5.5. Each post was annotated in a single labelling pass with the instruction-tuned Fanar-2-27B checkpoint [3].

### 4.3 Human Validation

We validate a random 1,000-post subsample against human annotation using the same taxonomy. Two annotators labelled the posts independently. They disagreed on 399 posts, which a third annotator adjudicated without seeing the model's label. The gold label of a post is the adjudicated label set; for posts on which the two annotators agreed, it is their shared label set. For 12 of the disputed posts the adjudicator could not settle on a label set; we label these *Other/Unclear*. All gold-based results below use the full 1,000 posts. Agreement between the two annotators is computed on their independent labels, before adjudication.

Agreement is computed per label as Cohen's $\kappa$ (presence/absence across the subsample) and at the post level as exact-set match and mean Jaccard overlap; per-label precision, recall, and $F_1$ take the gold labels as the reference (Table 1).

Overall agreement with the gold is substantial: macro $\kappa = 0.74$ (bootstrap 95% CI [0.71, 0.77]), in the "substantial" band of Landis and Koch (1977), with exact-set match of 68.2% and mean Jaccard of 0.79. Agreement is highest for *Religious Framing* ($\kappa = 0.97$), *Skeptical-Warning* ($\kappa = 0.80$), and *Market Commentary* ($\kappa = 0.78$), which are the three modes on which the model-generated annotation layer is well-calibrated. One systematic divergence stands out: the model *over-applies Promotional*, with high recall and low precision, assigning the label to 232 posts against 141 in the gold; *Educational* is over-applied to a lesser degree (249 against 218). *Other/Unclear*, by contrast, is applied somewhat less often than in the gold (61 against 77 posts) and agreement is moderate ($\kappa = 0.60$; precision 0.70, recall 0.56). We carry these biases forward as explicit prevalence bounds in Section 5.2 rather than leaving them implicit.

[1] CrowdTangle was a Meta public-insights tool used to track and analyze public Facebook content. It was discontinued on August 14, 2024, and replaced by the Meta Content Library and Content Library API.

[2] https://api.fanar.qa/

[3] The annotation prompt is reproduced in Appendix

Because $\kappa$ is unstable for rare labels, we also computed PABAK, which ranges from 0.76 to 1.00 across the six modes. Positive specific agreement is high for *Market Commentary* (0.91) and *Skeptical-Warning* (0.86) and lowest for *Other/Unclear* and *Promotional* (0.62 and 0.68); negative agreement is at least 0.87 throughout. *Religious Framing*, with 16 gold instances, retains a positive agreement of 0.97.

Agreement between the two annotators, before adjudication, is macro $\kappa = 0.62$ (exact-set match 60.3%; Table 1, last column), below Fanar's agreement with either of them (0.70 and 0.69), so the model's disagreements with an individual annotator lie within the range of disagreement between people. Model–gold exact-set match is 85.2% on the 601 posts on which the two annotators agreed and 43.9% on the 387 posts resolved by the third annotator, so the model's errors concentrate where people disagree.

All three annotators are native Arabic speakers familiar with the dialects in the corpus, and they received the same full guidelines as the model prompt (Appendix). The third annotator labelled the posts and categories on which the first two disagreed. Model-generated prevalence in the validation sample differs from the remaining 9,000 posts by at most 2.3 points per label, and median engagement is identical (8 vs. 8).

# 5 Results

## 5.1 Label Prevalence and Multi-functionality (RQ1, RQ2)

Table 2 reports per-label prevalence from raw model output. *Market Commentary* is the dominant function, accounting for 56.0% of posts, followed by *Skeptical-Warning* (31.1%). *Religious Framing* is rare at 1.7%, which is a direct consequence of the organic distribution that preserves but cannot statistically amplify infrequent functions. The *Other/Unclear* value of 10.5% is the raw rate; after resolving the exclusivity violations (Section 5.5) it falls to 5.6%, which is the figure used in all co-occurrence, engagement, and calibration analyses.

The discourse is substantially multi-functional. As shown in Table 3, 38.0% of posts carry two or more labels after the *Other/Unclear* repair (Section 5.5; 41.8% before it); two-label posts alone account for 35.0% of the corpus, and 3.0% carry three or more. On the 1,000 validation posts the annotators give two or more labels to 22.8% and 27.2% of posts, the gold to 29.6%, and the model to 39.8%.

## 5.2 Calibration-Adjusted Prevalence (RQ1)

Because validation (Table 1) reveals systematic over-application, we report calibration-adjusted prevalence using the Rogan-Gladen estimator, which corrects an observed proportion by the per-label sensitivity (recall) and specificity measured against the gold labels of the validation subset (Table 4). The adjustment revises *Promotional* (23.9 to 15.0%) substantially downward and *Educational* (23.3 to 19.5%) moderately downward, raises *Other/Unclear* (5.6 to 6.8%) and *Market Commentary* (56.0 to 57.1%) slightly, and leaves *Skeptical-Warning* and *Religious Framing* close to their raw values, which mirrors the agreement profile. The choice of reference matters. With the first annotator alone, *Educational* falls to 12.7% and *Other/Unclear* rises to 13.1%, because that annotator applied *Other/Unclear* to 144 posts against 77 in the gold; with the second annotator alone, *Educational* stays at 24.0% (Table 4). Bootstrap 95% intervals over the gold subset (Table 4) exclude the raw model point estimate for *Educational* and *Promotional*, confirming that the over-application is not an artefact of validation-sample noise; for the other modes the intervals include the raw value. The raw model prevalences for *Educational* and *Promotional* should thus be treated as upper bounds against the gold judgement.

## 5.3 Qualitative Error Analysis

Table 5 shows representative model-gold disagreements that explain the calibration biases. *Educational* over-application (74 posts that the gold does not label Educational, 47 of them Market Commentary in the gold) tracks surface cues, where posts containing تطوير ("development") or كيف ("how") are tagged *Educational* even when the gold reads market commentary. *Promotional* over-application (106 posts, 84 of them Market Commentary in the gold) is triggered by aspirational hype, e.g. قد تجعلك مليونيرًا ("could make you a millionaire"), that the gold treats as commentary rather than a call to act. *Other/Unclear* is applied less often than in the gold (61 against 77 posts; recall 0.56). The model misses 34 gold *Other/Unclear* posts and maps 12 of them onto *Skeptical-Warning* and 19 onto *Market Commentary*; the former typically occurs when the post reports a crypto-related harm, such as theft or fraud convictions, whereas the gold reserves *Skeptical-Warning* for posts that warn the *reader* directly and assigns *Other/Unclear* to third-party news events. The errors are cue-driven and systematic rather than random, which is precisely what makes the Rogan-Gladen correction meaningful.

Table 1: Model (Fanar) vs. gold agreement on the 1,000-post validation subsample (resolved labels). $n_g/n_s$: posts carrying the label in the gold/model annotation. P, R, $F_1$ take the gold labels as reference. $\kappa_{hh}$: agreement between the two annotators' independent labels before adjudication (all 1,000 posts).

| Mode | $n_g$ | $n_s$ | $\kappa$ | P | R | $F_1$ | $\kappa_{hh}$ |
|---|---|---|---|---|---|---|---|
| Religious Framing | 16 | 17 | 0.97 | 0.94 | 1.00 | 0.97 | 0.72 |
| Skeptical-Warning | 276 | 290 | 0.80 | 0.84 | 0.88 | 0.86 | 0.69 |
| Market Commentary | 597 | 580 | 0.78 | 0.92 | 0.90 | 0.91 | 0.65 |
| Educational | 218 | 249 | 0.67 | 0.70 | 0.80 | 0.75 | 0.49 |
| Promotional | 141 | 232 | 0.61 | 0.54 | 0.89 | 0.68 | 0.68 |
| Other/Unclear | 77 | 61 | 0.60 | 0.70 | 0.56 | 0.62 | 0.46 |
| Macro $\kappa/F_1$ | | | 0.74 | | | 0.80 | 0.62 |
| Exact-set match | | | | | | 68.2% | 60.3% |
| Mean Jaccard | | | | | | 0.79 | |

Table 2: Per-label prevalence over the 10,000-post sample from raw model output. Labels are not mutually exclusive, so column sums exceed 100%.

| Communicative mode | % of corpus |
|---|---|
| Market Commentary | 56.0 |
| Skeptical-Warning | 31.1 |
| Promotional | 23.9 |
| Educational | 23.3 |
| Other/Unclear | 10.5 |
| Religious Framing | 1.7 |

Table 3: Distribution of labels per post (resolved labels).

| Labels per post | Posts | % |
|---|---|---|
| 1 | 6,203 | 62.0 |
| 2 | 3,502 | 35.0 |
| 3 | 229 | 2.3 |
| 4 | 66 | 0.7 |
| Multi-label (≥ 2) | 3,797 | 38.0 |

### 5.4 Co-occurrence Structure (RQ2)

Table 6 reports pairwise co-occurrence among the five substantive modes, with *Other/Unclear* excluded by construction, together with the count expected if the two labels were assigned independently (i.e. $E = N\, p_A\, p_B$, where $p$ denotes the per-label prevalences of Table 2) and the resulting enrichment (lift = observed/expected; lift > 1 indicates over-representation relative to chance). Reading raw frequency and enrichment together is essential here because the two diverge.

In absolute terms the leading pair is *Market Commentary* with *Skeptical-Warning* at 14.7% of the corpus (1,471 posts), but this is largely a by-product of the two labels' high prevalence: against the independence baseline of 17.4% the pair is *not* enriched (lift 0.85). Caution is thus distributed broadly across market discourse rather than concentrated in a separate warning genre.

The single pair that is positively enriched is *Educational* with *Promotional* (8.8% observed vs. 5.6% expected; lift 1.58), where instructional content and calls to act co-occur markedly more than chance would predict. This suggests a "teach-then-sell" tendency in which explanation may serve as an on-ramp to advocacy, though because the labels are unordered the directional reading is interpretive rather than established. One caveat tempers this reading: the model over-applies *Promotional* and, to a lesser degree, *Educational* (Table 1), and the error analysis shows this over-application is cue-driven on overlapping promotional content, meaning part of the observed enrichment may reflect correlated over-tagging rather than a property of the discourse. On human labels the pairing holds in the gold (lift 1.59, 49 posts), for the first annotator (2.26) and on posts where both annotators agree (1.80, 16 posts), not for the second (1.02). Every other pair sits below its independence baseline (lift $<$ 1), most strongly *Promotional* with *Skeptical-Warning* (0.25) and *Educational* with *Skeptical-Warning* (0.38), indicating that posts that warn tend not to also sell or teach. *Religious Framing* rarely co-occurs with any mode, and it never co-occurs with *Market Commentary*: none of the 10,000 posts carries both labels (95% Wilson interval [0.00, 0.04], against 1.0% expected). Given the corpus size, 95% Wilson intervals on the observed proportions are tight, e.g. *Market Commentary* with *Skeptical-Warning* [14.0, 15.4] and *Educational* with *Promotional* [8.3, 9.4].

Table 4: Raw model output vs. Rogan-Gladen calibration-adjusted prevalence (resolved labels; sensitivity/specificity from the 1,000 validation posts, Table 1). 95% CI from 3,000 bootstrap resamples of the gold subset. Adj. (1st) and Adj. (2nd) repeat the adjustment with the first and the second annotator as reference.

| Mode | Raw % | Adj. % | 95% CI | Adj. (1st) % | Adj. (2nd) % |
|---|---|---|---|---|---|
| Market Commentary | 56.0 | 57.1 | [54.6, 59.6] | 55.4 | 51.3 |
| Skeptical-Warning | 31.1 | 30.2 | [28.1, 32.2] | 25.1 | 28.2 |
| Educational | 23.3 | 19.5 | [16.8, 22.4] | 12.7 | 24.0 |
| Promotional | 23.9 | 15.0 | [12.2, 17.6] | 13.2 | 16.0 |
| Other/Unclear | 5.6 | 6.8 | [4.9, 9.3] | 13.1 | 5.2 |
| Religious Framing | 1.7 | 1.6 | [1.4, 1.7] | 1.3 | 0.9 |

Table 5: Representative model–gold disagreements from the 1,000-post validation subsample, with English translations. Errors are cue-driven and consistent with the calibration biases.

| Post (excerpt) | Translation | Model | Gold | Pattern |
|---|---|---|---|---|
| تطوير العملة الرقمية وليس المشفرة | Developing the digital currency, not the cryptocurrency | Educational | Market Commentary | Educational over-applied on "development" cue |
| عملة البتكوين قد تجعلك مليونيراً بدون مجهود | Bitcoin could make you a millionaire without effort | Promotional | Market Commentary | Promotional over-applied on hype; no call to act |
| بعد سرقة العملات... تعويضات بالملايين | After the theft of the coins… compensation worth millions | Skeptical-Warning | Other/Unclear | Other/Unclear missed; third-party news read as warning |

### 5.5 Annotation Quality and Constraint Adherence

On raw output, before any constraint repair, prompt-only control was high but not perfect. Zero out-of-taxonomy labels were produced. The *Other/Unclear* exclusivity constraint was honoured in 95.14% of posts; 486 posts (4.86% of the corpus, which represents 46.3% of the 1,049 raw *Other/Unclear*-tagged posts) attached *Other/Unclear* alongside a substantive label. A one-line deterministic rule, specifically dropping *Other/Unclear* whenever a substantive label is present, eliminates all 486 violations. We apply exactly this rule to obtain the resolved labels used throughout and report the pre-resolution rate as a transparency measure of prompt-only adherence rather than as residual error. This leaves 563 genuinely residual posts (5.6%).

### 5.6 Engagement by Communicative Mode

Table 7 relates communicative mode to engagement. We report both mean and median Total Interactions since the distribution is strongly right-skewed (corpus mean 119.6, median 8), and engagement is computed directly on the annotated corpus. *Religious Framing* shows the highest mean (289.1) but rests on only 171 posts, so it is reported with caution. Multi-label posts show slightly higher engagement than single-label posts (mean 132.9 vs. 111.4; median 9 vs. 8), but a bootstrap of the difference does not exclude zero (mean difference 21.4, 95% CI [−0.4, 44.8]; median difference 1, 95% CI [0, 3]; 3,000 resamples), so we treat this as at most a weak association. All associations reported here are descriptive and correlational; they do not establish that communicative mode *causes* engagement, which is plausibly confounded by page size, topic salience, and posting time.

Table 6: Pairwise co-occurrence among substantive modes (resolved labels), ordered by enrichment. Obs: observed % of 10,000 posts. Exp: % expected if the two labels were assigned independently (i.e. $E = N p_A p_B$). Lift = Obs/Exp; lift > 1 is over-, < 1 under-representation relative to chance.

| Mode pair | Obs % | Exp % | Lift |
|---|---|---|---|
| Educational + Promotional | 8.8 | 5.6 | 1.58 |
| Market Commentary + Skeptical-Warning | 14.7 | 17.4 | 0.85 |
| Market Commentary + Promotional | 9.8 | 13.4 | 0.74 |
| Educational + Market Commentary | 7.3 | 13.0 | 0.56 |
| Religious Framing + Skeptical-Warning | 0.2 | 0.5 | 0.45 |
| Promotional + Religious Framing | 0.2 | 0.4 | 0.42 |
| Educational + Religious Framing | 0.2 | 0.4 | 0.40 |
| Educational + Skeptical-Warning | 2.8 | 7.2 | 0.38 |
| Promotional + Skeptical-Warning | 1.9 | 7.4 | 0.25 |
| Market Commentary + Religious Framing | 0.0 | 1.0 | 0.00 |

Table 7: Engagement by communicative mode (resolved labels). Means and medians of Total Interactions; corpus row for reference.

| Mode | Posts | Mean | Median |
|---|---|---|---|
| Religious Framing | 171 | 289.1 | 19 |
| Educational | 2,330 | 143.9 | 12 |
| Skeptical-Warning | 3,108 | 129.6 | 10 |
| Promotional | 2,388 | 125.7 | 9 |
| Other/Unclear | 563 | 125.0 | 8 |
| Market Commentary | 5,598 | 104.6 | 6 |
| Single-label posts | 6,203 | 111.4 | 8 |
| Multi-label posts | 3,797 | 132.9 | 9 |
| Corpus | 10,000 | 119.6 | 8 |

## 5.7 Reaction Profiles by Communicative Function

Reaction profiles differ clearly by communicative function: Love dominates Educational and Promotional posts, Skeptical-Warning draws the largest Haha share, and Market Commentary the largest Wow share. Table 8 reports the mean per-post share of each reaction by mode. We use single-label posts from 2020 onward that received at least one of the six reactions (1,857 of the 10,000 posts) and report page-clustered bootstrap intervals. Reaction profiles differ clearly by mode. Love is the largest share for Educational (71.3%) and Promotional posts (68.4%) and the smallest for Skeptical-Warning (33.3%); the Love gap between Educational and Skeptical-Warning is 38.0 points (95% CI [31.4, 44.6]) and between Promotional and Skeptical-Warning 35.1 points ([28.6, 41.3]). Skeptical-Warning draws the largest Haha share (39.0%), 20.5 points above Market Commentary ([15.4, 26.4]). Market Commentary has the highest Wow share (19.7%), well above Educational (7.0%) and Promotional (4.1%). These patterns hold when the ten pages with the most reactions are removed, when posts with fewer than five reactions are excluded, and when multi-label posts are included. Religious Framing has only 53 such posts and its estimates are imprecise, so we do not interpret them. Reactions are a behavioural proxy for audience response and can encode approval or stance rather than affect (Tardelli et al., 2026); we therefore describe the profiles without reading Haha as irony.

## 5.8 Temporal Shift in Functional Composition (RQ4)

Splitting the corpus at 2020 (Table 9), the share of posts carrying *Market Commentary* rises from 46.3% in 2014-2019 to 57.7% in 2020-2024, while *Educational* falls from 31.6% to 21.8% and *Promotional* rises from 20.5% to 24.5%. *Skeptical-Warning* stays broadly stable, moving from 29.5% to 31.4%. The shift is consistent with a maturing discourse in which foundational explanation gives way to price talk and promotion over a period whose post volume is itself concentrated in 2021-2022. Because the two windows differ greatly in size, we present this as a descriptive composition shift rather than a controlled trend.

# 6 Discussion

The most frequent pairing, *Market Commentary* with *Skeptical-Warning*, mostly reflects the two labels' high prevalence (lift 0.85), so caution is spread across market discourse rather than concentrated in a separate warning genre. The pairing most over-represented relative to chance is *Educational* with *Promotional* (lift 1.58), a possible "teach-then-sell" pattern. The model over-applies both labels (Table 1), so part of this enrichment may be correlated over-tagging, but the gold labels

Table 8: Reaction composition by communicative mode: mean per-post share (%) of each of the six reactions, with 95% page-clustered bootstrap intervals for Love, Wow and Haha. Single-label posts from 2020 onward with at least one of the six reactions ($n$ = 1,857). Sad, Angry and Care are point estimates. Reactions are a behavioural proxy, not a validated measure of affect.

| Mode | Posts | Pages | Love | Wow | Haha | Sad | Angry | Care |
|---|---|---|---|---|---|---|---|---|
| *Market Commentary* | 691 | 256 | 50.5 [46.1, 56.2] | 19.7 [14.1, 25.4] | 18.4 [14.2, 22.5] | 3.0 | 1.5 | 6.9 |
| *Educational* | 217 | 169 | 71.3 [65.8, 76.2] | 7.0 [3.9, 10.3] | 9.8 [6.8, 13.2] | 1.4 | 2.4 | 8.1 |
| *Promotional* | 211 | 148 | 68.4 [63.5, 73.2] | 4.1 [1.8, 6.6] | 13.9 [9.7, 18.8] | 1.7 | 3.2 | 8.8 |
| *Skeptical-Warning* | 478 | 304 | 33.3 [30.0, 36.9] | 14.6 [12.2, 17.2] | 39.0 [35.3, 42.9] | 2.4 | 3.8 | 6.9 |
| *Religious Framing* | 53 | 51 | 58.4 [47.3, 69.4] | 8.7 [2.9, 16.4] | 24.1 [14.7, 33.5] | 2.3 | 0.2 | 6.3 |
| *Other/Unclear* | 207 | 179 | 50.3 [44.0, 56.5] | 7.3 [4.7, 10.2] | 24.5 [18.6, 30.7] | 5.5 | 3.5 | 8.9 |

Table 9: Mode prevalence by era (% of posts in each window carrying the mode).

| Mode | 2014-2019 ($n$=1,499) | 2020-2024 ($n$=8,501) |
|---|---|---|
| *Market Commentary* | 46.3 | 57.7 |
| *Educational* | 31.6 | 21.8 |
| *Promotional* | 20.5 | 24.5 |
| *Skeptical-Warning* | 29.5 | 31.4 |
| *Religious Framing* | 2.3 | 1.6 |

show the same enrichment (lift 1.59), which argues against a pure artefact.

Reaction profiles separate the modes into three groups: Love-dominated (Educational, Promotional), mixed (Market Commentary) and Haha-leaning (Skeptical-Warning). Because reactions track what a post is doing, they cannot be read as a direct measure of how audiences feel. We do not control for the sentiment of the text, so we cannot say how much of the difference reflects function rather than feeling.

**Intended use and scope.** AraCrypto-ML supports three uses. First, it can be used to train and evaluate classifiers of communicative function in Arabic financial discourse: the model labels for the 10,000 posts form a silver training layer, and the 1,000-post subset with adjudicated gold labels is a test set (we recommend page-grouped splits, since posts from one page are not independent). Second, the multi-label design lets researchers study how functions combine; for example, *Promotional* posts rarely also carry *Skeptical-Warning* (lift [0.25]). Third, because reaction profiles differ by function, the labels let researchers separate what a post does from how audiences react to it, which matters for work that uses reactions as a proxy for affect. Raw model prevalences should not be used as population estimates: *Promotional* and, to a lesser degree, *Educational* are over-assigned, and *Religious Framing* has too few posts for reliable estimates.

## Data Availability

We will release the AraCrypto-ML corpus of 10,000 posts with their Fanar-2-27B labels and the human annotations and adjudicated gold labels for the 1,000-post validation subset. The data will be available through an access form.

## 7 Conclusion

We introduced **AraCrypto-ML**, a 10,000-post multi-label corpus of communicative functions in Arabic Facebook cryptocurrency discourse, drawn from 2,898 public pages over a decade. Fanar-2-27B labels were validated against a 1,000-post subsample with adjudicated human gold labels (macro $\kappa$ = 0.74, against 0.62 between the two annotators before adjudication) and calibrated with the Rogan–Gladen estimator, so that adjusted prevalence bounds are reported alongside raw counts.

The discourse is highly multi-functional. The most frequent pairing, Market Commentary with Skeptical-Warning, mostly reflects the two labels' high prevalence, while Educational with Promotional is the only pair enriched beyond chance (lift 1.58). Reaction profiles differ by function: Love is about 70% of reactions to Educational and Promotional posts but about 33% for Skeptical-Warning, which draws the largest Haha share. After 2020, the composition shifts from explanation toward price talk and promotion. These patterns show structure that sentiment analysis alone does not capture. We release AraCrypto-ML as an annotated resource rather than a benchmark, since we report no classifier baselines, for use in Arabic NLP, computational social science, and misinformation detection in digital financial contexts.

## Limitations

Annotations are produced by a single model (Fanar-2-27B) in a single labelling pass under a fixed prompt, without decoding hyperparameter sweeps or multi-sample self-consistency checks. Validation against the gold labels of a 1,000-post subsample yields substantial overall agreement (macro $\kappa = 0.74$) but reveals systematic over-application of *Promotional* and, to a lesser degree, *Educational*, which we address through Rogan-Gladen calibration rather than leaving implicit. The two annotators agree with each other at macro $\kappa = 0.62$ before adjudication, and calibrating against either annotator alone gives different corrections for *Educational* and *Other/Unclear* (Section 5.2). The adjudicator saw the two annotators' labels, so the gold labels for the disputed posts are not independent of them. Statistical power is weak for rare modes, notably *Religious Framing* (1.7%, 171 posts), so its engagement and reaction profiles are indicative only. Reaction analyses use single-label posts from 2020 onward with at least one of the six reactions (1,857 posts); Religious Framing has too few such posts (53) to interpret. We do not control for the sentiment of the post text, so we cannot say how much of the reaction differences reflects function rather than feeling. AraCrypto-ML is released as an annotated resource rather than a benchmark: we train and evaluate no classifiers on it, so task difficulty and model rankings on these labels remain untested.

Collection is keyword-based and may under-capture Arabizi, Latin-script tickers, and dialectal spellings. Engagement and reaction relationships are reported without regression controls for page size, posting time, or geography and should be read as descriptive and correlational rather than causal. Co-occurrence lift values are computed on uncalibrated model-generated labels against a first-order independence baseline and should be read as descriptive associations rather than evidence of authorial intent. The corpus covers public pages only, and pretraining contamination between Fanar-2 and the target pages cannot be excluded.

Future work will train and evaluate classifier baselines, including fine-tuned encoders (Berriche et al., 2025), on page-grouped splits, compare Fanar against alternative annotators, and model engagement by mode with regression controls for page, time, and geography.

## Ethical Considerations

This study analyses publicly available Facebook posts and reports only aggregate results; the post excerpts shown in the paper carry no author or page identifiers. It follows established ethical guidelines for privacy protection and harm minimisation in computational social science. Annotation was performed with an Arabic-centric model (Fanar) developed with attention to Arabic cultural and linguistic context; we treat its labels as provisional pending broader human validation.

The human validation study involved three bilingual annotators recruited from the research team's institutional network. The annotators were informed of the nature and purpose of the task prior to participation, provided consent to take part, and were compensated for their time. No personally identifiable information about the annotators is reported in this work. The study protocol was reviewed and approved by the authors' institutional review board (IRB).

## AI Assistance

AI tools were used during the preparation of this work in two capacities. First, Fanar-2-27B was used as the annotation instrument for the corpus, under a prompt-only multi-label protocol as described in Section 4.2; its outputs are treated as model-generated labels subject to human validation and calibration as reported in Section 4.3. Second, AI writing assistants were used for improvements to grammar and clarity during manuscript preparation. All AI-generated text was carefully reviewed and validated by the authors. The authors take full responsibility for the accuracy, originality, and integrity of the manuscript and analyses.

## Appendix

We reproduce the operative content of the multi-label prompt below, specifically the pre-classification checklist, the labelling constraints, and the forbidden-label mappings as given to the model. Per-label definitions are as in Table A1. Labels appear in the prompt in their full canonical form (e.g. `Market / Price Commentary`); the abbreviated forms used throughout the paper (e.g. *Market Commentary*) refer to the same categories without change of meaning. Repetitive formatting instructions are omitted for space.

**Objective.** Annotate each Arabic Facebook cryptocurrency post by identifying *all* communicative

Table A1: Communicative-mode taxonomy. Cue terms are illustrative, not exhaustive. *Other/Unclear* is mutually exclusive with all substantive labels by definition.

| Label | Definition | Illustrative Arabic cues |
|---|---|---|
| Market Commentary | Reports, analyses, predicts, or comments on prices, markets, trading, or investment performance; observational or analytical in tone. | السعر (price), ارتفاع (increase), تحليل فني (technical analysis), تداول (trading), السوق (market) |
| *Educational* | Explains concepts, technologies, or procedures; instructional rather than persuasive or risk-alerting. | ما هو (what is), كيف (how), شرح (explanation), بلوكشين (blockchain), محفظة (wallet) |
| *Promotional* | Persuades the reader to act, e.g. join, sign up, invest, or use a referral; defined by a call to action. | انضم (join), سجل الآن (register now), فرصة (opportunity), رابط التسجيل (registration link) |
| *Skeptical-Warning* | Alerts the reader to danger, that is, fraud, hacks, crashes, or bans, with caution as the dominant intent. Also applies to government or regulatory actions, such as bans, crackdowns, or central-bank warnings, when framed as risk, restriction, or legitimacy concern rather than neutral reporting. | احتيال (fraud), تحذير (warning), خطر (risk), فقاعة (bubble), حظر (ban) |
| *Religious Framing* | Evaluates cryptocurrency through Islamic jurisprudence (Shariah), including halal/haram judgements, fatwas, and discussions of gharar, maysir, or riba. Strictly religious in scope; state regulation, government bans, and ordinary legal or policy actions are not included unless the post explicitly connects them to Islamic/Shariah rulings. | حلال (halal), حرام (haram), فتوى (fatwa), غرر (gharar), الشريعة (Shariah) |
| *Other/Unclear* | Applies when none of the five substantive functions apply, including off-topic, purely personal, or non-crypto content. Never combined with another label. | – |

categories that meaningfully apply. A post may receive one or more labels.

**Pre-classification checklist (evaluate independently).** (1) Does it discuss prices, markets, trading, or investment performance? (2) Does it explain or teach how something works? (3) Does it promote, sell, or call readers to act? (4) Does it warn readers about danger, fraud, risk, or harm? (5) Does it evaluate cryptocurrency through an Islamic/Shariah lens? Assign the corresponding label for every "yes"; if none, assign `Other/Unclear`.

**Constraints.** Use only the six canonical labels; never invent, rename, or merge labels. `Other/Unclear` is assigned only when none of labels 1-5 apply and must never co-occur with another label. Output only a JSON array of label strings, with no explanation.

**Forbidden-label mappings.** Out-of-taxonomy labels should be mapped as follows: Regulatory News, Compliance, or Regulatory Framing → `Skeptical/Warning`; News Reporting, Environmental Impact, Historical/Commemorative → `Other/Unclear`.

**Prompt:** We ran Fanar-2-27B through the Fanar API on 7 June 2026 with temperature 0, using the following prompt. **Role and task.** You are annotating Arabic Facebook posts about cryptocurrency and digital-currency discourse. Task: assign all semantic labels that meaningfully apply to the post. Output only a valid JSON array. No explanation.

**Allowed labels.** `Market / Price Commentary`; `Educational / Explanatory`; `Promotional / Advocacy`; `Skeptical / Warning`; `Religious / Legal Framing`; `Other / Unclear`.

**Market / Price Commentary.** Use for posts discussing specific cryptocurrencies, price movements, market predictions, trading observations, gains/losses, investment performance, or technical/chart analysis. Arabic cues: البتكوين، البيتكوين، سعر، أسعار، ارتفاع، انخفاض، صعود، هبوط، تداول، صفقة، ربح، خسارة، أرباح، خسائر، تحليل فني، شارت، دعم، مقاومة، هدف سعري

**Educational / Explanatory.** Use for posts explaining how crypto, blockchain, wallets, exchanges, mining, or related technologies work. Also use for onboarding tutorials, definitions, guides, and how-to content. Arabic cues: كيفية، كيف، شرح، ما

هو، تعلم، دليل، خطوات، محفظة، بلوكشين، سلسلة الكتل، منصة تداول، تعدين، عقد ذكي، تعريف

**Promotional / Advocacy.** Use for posts promoting a specific exchange, token, signal group, investment opportunity, or crypto adoption. Also use when the post calls readers to buy, invest, join, register, subscribe, or use a referral link/code. Arabic cues: انضم، اشترك، سجل الآن، سجل، استثمر، اشترِ، فرصة، لا تفوت، أفضل منصة، رابط التسجيل، رابط الإحالة، استخدم الكود، توصيات، جروب إشارات، اربح

**Skeptical / Warning.** Use for posts warning about scams, fraud, hacks, market crashes, bubbles, losses, or dangerous speculation. Also use for posts criticizing crypto hype or sarcastically engaging with unrealistic promises. Use this label for government or regulatory actions when framed as risk, caution, restriction, or legitimacy concern, such as bans, crackdowns, central-bank warnings, compliance requirements, or legal/policy restrictions. Key test: is the post trying to make readers cautious, doubtful, or protect them from harm? Arabic cues: احتيال، نصب، خداع، تحذير، احذروا، احذر، خطر، فقاعة، انهيار، اختراق، خسارة، حظر، منع، تضييق، البنك المركزي، تنظيم، قانون، قيود، مضاربة، وهم، تجنب

**Religious / Legal Framing.** Use only when cryptocurrency is evaluated through Islamic/Shariah framing, such as halal/haram judgments, fatwas, fiqh discussions, permissibility, gharar, maysir, riba, or moral/religious arguments about crypto. This label is strictly religious/Shariah-based. Do not use it for state regulation, government bans, or ordinary legal/policy actions unless the post explicitly connects them to Islamic/Shariah rulings. Arabic cues: حلال، حرام، فتوى، حكم الشرع، حكم شرعي، جائز، غير جائز، شرعي، الشريعة، الفقه، ميسر، غرر، ربا

**Other / Unclear.** Use only when none of the above labels apply. Includes: off-topic posts, non-crypto posts, purely personal updates, generic news with no crypto-specific angle, environmental/energy commentary without a market/educational/promotional/warning/religious angle, historical or commemorative posts, or unclear content.

**Rules.** (1) Assign every applicable label. (2) A post may receive multiple labels. (3) Use only the allowed labels exactly as written. (4) Do not invent, rename, split, merge, or add labels. (5) Do not output duplicate labels. (6) Keywords are illustrative cues only; rely on the overall meaning and intent of the post. (7) Neutral market-impact reporting may be `Market / Price Commentary`. (8) Neutral government/regulatory news is not `Religious / Legal Framing` unless it explicitly uses Islamic/Shariah framing. (9) Regulatory actions, bans, restrictions, or central-bank warnings framed as risk or caution should be `Skeptical / Warning`. (10) `Other / Unclear` must be used only when none of the other labels apply. (11) `Other / Unclear` must never appear with any other label.